\pdfoutput=1

\documentclass[11pt]{article}

\usepackage{EMNLP2023}

\usepackage{times}
\usepackage{latexsym}

\usepackage[T1]{fontenc}

\usepackage[utf8]{inputenc}

\usepackage{microtype}

\usepackage{inconsolata}

\usepackage{amsmath}
\usepackage{graphicx}
\usepackage{booktabs}
\usepackage{amssymb}
\usepackage{transparent}
\usepackage{multirow}
\usepackage{tikz}
\usepackage[capitalize]{cleveref}
\usepackage{pifont}
\usepackage[super]{nth}
\usepackage[shortlabels]{enumitem}
\usepackage{quoting}
\usepackage{subcaption}
\usepackage{adjustbox}
\usepackage[ruled,vlined]{algorithm2e}

\usepackage{array}
\usepackage{booktabs}
\usepackage{tablefootnote}

\usepackage{fancyvrb} 
\VerbatimFootnotes    

\crefformat{section}{\S#2#1#3} 
\crefformat{subsection}{\S#2#1#3}
\crefformat{subsubsection}{\S#2#1#3}

\mathchardef\mhyphen="2D 

\title{Neural Architecture Search for Effective Teacher-Student Knowledge Transfer in Language Models}

\author{Aashka Trivedi, Takuma Udagawa, Michele Merler, Rameswar Panda, \\ {\bf Yousef El-Kurdi, Bishwaranjan Bhattacharjee} \\
IBM Research AI \\
\texttt{\{aashka.trivedi@, takuma.udagawa@, mimerler@us., rpanda@ }\\
\texttt{yousefelk@us., bhatta@us.\}ibm.com}}

\begin{document}
\maketitle
\begin{abstract}
Large pretrained language models have achieved state-of-the-art results on a variety of downstream tasks. Knowledge Distillation (KD) into a smaller student model addresses their \emph{inefficiency}, allowing for deployment in resource-constrained environments. However, KD can be \emph{ineffective} when the student is manually selected from a set of existing options, since it can be a sub-optimal choice within the space of all possible student architectures. We develop multilingual KD-NAS, the use of Neural Architecture Search (NAS) guided by KD to find the optimal student architecture for task agnostic distillation from a multilingual teacher. In each episode of the search process, a NAS controller predicts a reward based on the distillation loss and latency of inference. The top candidate architectures are then distilled from the teacher on a small proxy set. Finally the architecture(s) with the highest reward is selected, and distilled on the full training corpus. KD-NAS can automatically trade off efficiency and effectiveness, and recommends architectures suitable to various latency budgets. Using our multi-layer hidden state distillation process, our KD-NAS student model achieves a 7x speedup on CPU inference (2x on GPU) compared to a XLM-Roberta Base Teacher, while maintaining 90\% performance, and has been deployed in 3 software offerings requiring large throughput, low latency and deployment on CPU.

\end{abstract}

\begin{figure}[t!]
\includegraphics[width=\linewidth]{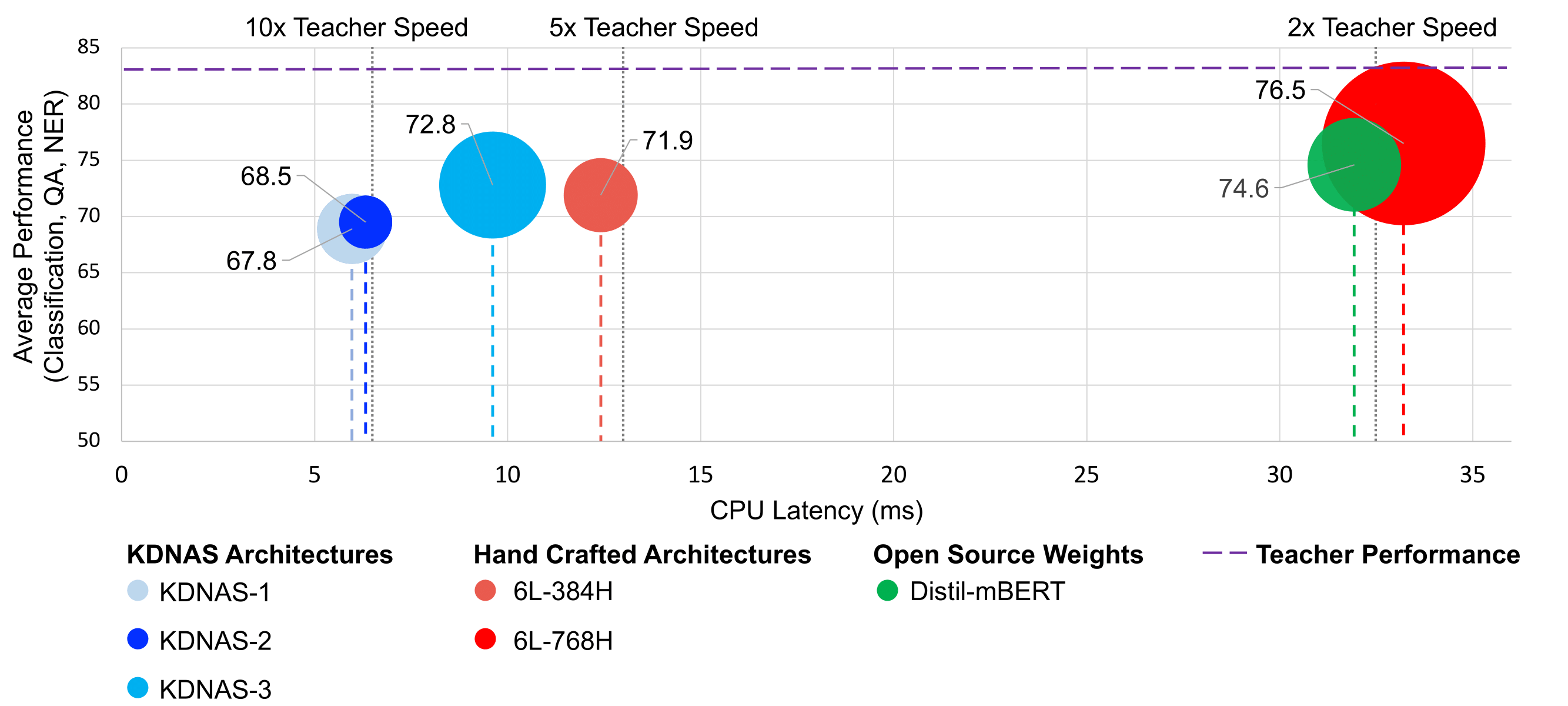}
\caption{KD-NAS architectures compared to hand-crafted baselines (bubble size indicates total parameters). At similar performance, KD-NAS selects faster architectures. At similar latency, architectures distilled on our pipeline outperform open-source baselines.}
\label{fig-models}
\end{figure}  

\begin{figure*}[t!]
\centering
\includegraphics[width=0.91\textwidth]{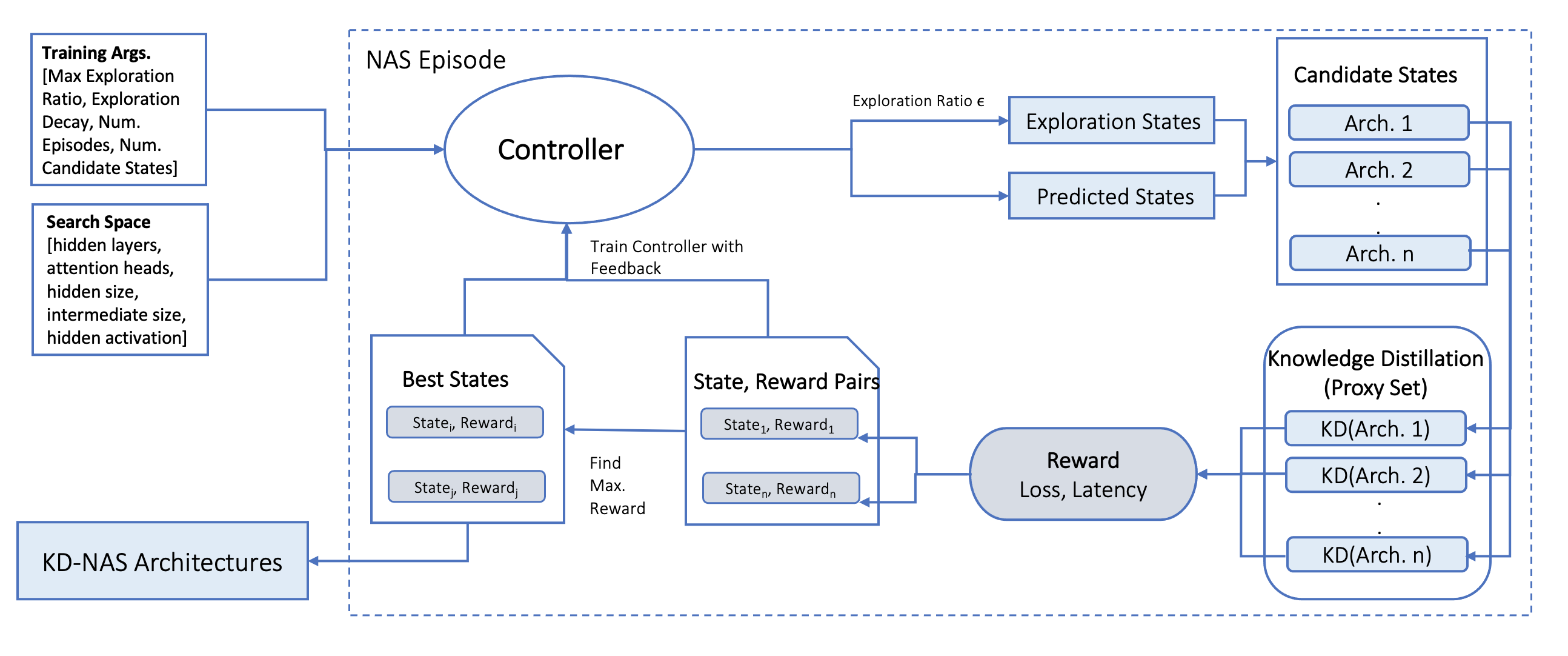}
\caption{KD-NAS. In each episode, the controller predicts the states with the highest reward. These states are chosen as candidate architectures to be distilled from the teacher in a reduced-KD process. The actual reward is obtained, and is used to train the controller, along with a memory of previously explored high performing states.}
\label{fig-kdnas}
\end{figure*}  

\section{Introduction}
Pretrained language models \cite{devlin2019bert,liu2019roberta, clark2020electra, conneau2020unsupervised} are capable of showing extraordinary performance on a variety of downstream language tasks. However, such models have a large number of parameters \cite{vaswani2017attention, devlin2019bert, liu2019roberta, brown2020language, kaplan2020scaling,fedus2021switch}, requiring huge amounts of memory to be deployed. Additionally, they have a large latency for inference - making them \emph{inefficient} to deploy in resource-constrained environments, or in products that require a large throughput, low latency and low memory footprint. Smaller models may be more suitable for use in practice, but often sacrifice performance for the sake of deployment \cite{huang2015bidirectional}. Moreover, these models are \emph{ineffective} due to their hand-crafted architecture design - in the space of all possible architectures given a certain resource budget, manual selection is likely sub-optimal even if guided by human expertise, and selecting the optimal architecture through trial-and-error for various latency budgets can be prohibitively expensive.

Knowledge Distillation (KD) \cite{hinton2015distilling} trains a smaller student model to mimic a larger teacher model. This improves the efficiency of current models, making the compressed model suitable for deployment without losing much performance. In practice, having a manually designed student with layers initialized from the teacher can be immensely beneficial to the distillation process \cite{wang-etal-2023-distill, sanh2020distilbert, turc2019pretrain}, but this puts a constraint on the architecture of the student. However, an optimal student architecture may exist, given a specific teacher model \cite{liu2020search}. This motivates us to focus on \emph{architecture-agnostic} distillation methods for student models which can have different architectural parameters from the teacher. Specifically, we follow a \emph{task-agnostic}\footnote{Task-agnostic distillation produces a general-purpose language model that can be finetuned on downstream tasks.} hidden state distillation objective \cite{jiao2020tinybert, jiao2021improving, mukherjee2021xtremedistiltransformers, ko2023revisiting}, and improve the performance of a multilingual student model \emph{without} pre-training or initialization from the teacher. Moreover, we use Neural Architecture Search (NAS) \cite{elsken2019neural, zoph2017neural, liu2019darts,so2019evolved} to efficiently automate the process of finding the best transformer architecture optimized for distilling knowledge from a multilingual teacher, obtaining faster architectures for a given performance level (\cref{fig-models}). Our key contributions are:
\begin{itemize}[topsep=0pt, itemsep=0pt, leftmargin=.15in, parsep=0pt]
    \item We develop KD-NAS, a system that improves the NAS process to identify an optimal yet efficient architecture from a pre-defined search space for a general purpose modelling objective. KD-NAS incorporates a KD objective with the traditional loss and latency measures into the NAS Reward. We introduce a Feedback Controller to guide the search process, which draws on information from the top performing states explored in previous episodes.
    \item We propose a multi-layer hidden state distillation approach, in which each student layer learns the internal representations of two teacher layers, allowing for the transfer of both low-level and high-level knowledge. This method effectively transfers knowledge without pre-training or weight initialization.
    \item We demonstrate how KD-NAS recommended architectures have been deployed for low-latency applications, achieving significant speedup while maintaining 90\% of teacher performance. 
\end{itemize}

\section{Background}

\paragraph{Knowledge Distillation} KD is a method of model compression in which the knowledge of a large model is transferred to a student through transferring its internal representations or outputs \cite{ba_2014,hinton2015distilling}. In transformer language models \cite{vaswani2017attention}, prior work has distilled teacher outputs \cite{hinton2015distilling,sanh2020distilbert, jafari-etal-2021-annealing}, hidden states \cite{mukherjee2021xtremedistiltransformers, sun-etal-2019-patient, jiao2021improving, li-etal-2020-bert}, multi-head self-attentions \cite{zagoruyko2017paying,wang2020minilm, wang-etal-2021-minilmv2}, or a combination of the above \cite{sanh2020distilbert, mukherjee2021xtremedistiltransformers}.  

This work adopts architecture-agnostic Hidden State (HS) distillation methods based on linear projections \cite{mukherjee2021xtremedistiltransformers,jiao2020tinybert}, in which a student is trained to learn the representations of the teacher's hidden state. Formally, for models with $L$ Transformer layers, the hidden state of the $i^{th}$ layer is $\mathbf{H}_i \in \mathbb{R}^{|x| \times d_h}$, where $d_h$ is the hidden size and $|x|$ is the sequence length. For HS Distillation from a teacher with $L^T$ layers to a student with $L^S$ layers (with hidden size ${d}^T_{h}$ and ${d}^S_{h}$ respectively), we map the $i^{th}$ student layer to a set of teacher layers, $g(i)$. For each $j \in g(i)$, we linearly transform the hidden state of the student $\mathbf{H}^S_{j} \in \mathbb{R}^{|x| \times {d}^S_{h}}$ using a learnable projection matrix $\mathbf{W}^j_{i} \in \mathbb{R}^{{d}^S_{h} \times {d}^T_{h}}$, and train the student to predict the hidden state of the teacher $\mathbf{H}^T_{j} \in \mathbb{R}^{|x| \times {d}^T_{h}}$ using the Mean Squared Error (MSE) loss:
\begin{equation}
\mathcal{L}_{\mathrm{HS}}(S,T) = \sum_{i=1}^{L^S} \sum_{j \in g(i)} \mathrm{MSE} \Bigl( \mathbf{H}^S_{j} \mathbf{W}_i^j, \mathbf{H}^T_{j} \Bigl)
\label{eq:hs_transfer_1_to_n}
\end{equation}


\begin{figure}[t!]
\centering
\includegraphics[width=0.4\textwidth]{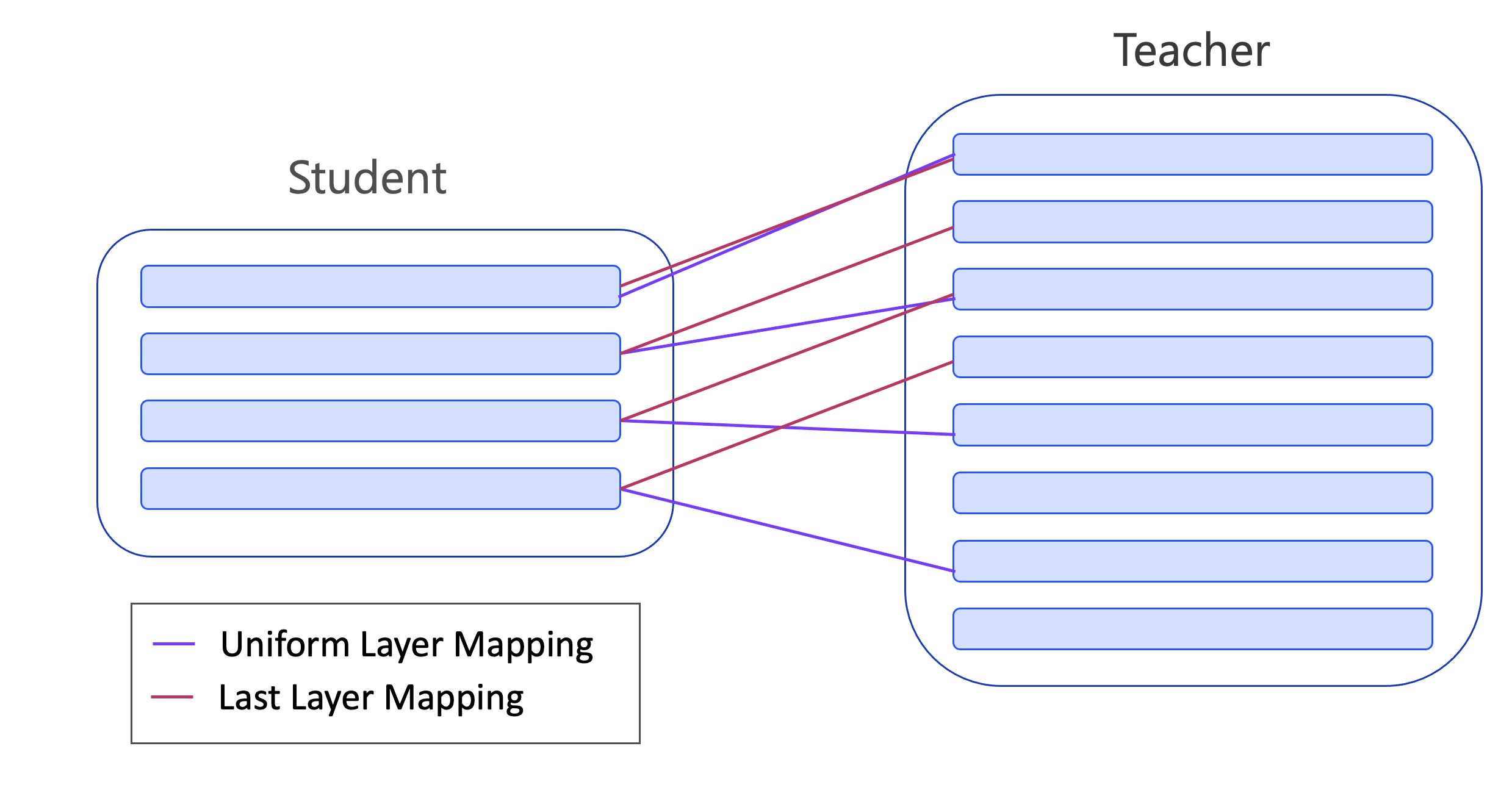}
\caption{Multi-layer mapping for HS Transfer. Lines from teacher layers to a student layer $i$ graphically depict the mapping function $g(i)$.}
\label{fig-distil}
\end{figure}  

This work proposes a \textit{multi-layer mapping strategy} \cite{jiao2021improving,wu-etal-2021-universal}, where each student layer learns from two teacher layers, as illustrated in \cref{fig-distil}. This allows the student to learn both  high-level (e.g. semantic) and low level (e.g. syntactic) knowledge in the teacher representations \cite{tenney2019bert}.  Specifically we set  $|g(i)|= 2 $ for $i \in [1, L^S]$, and combine the Uniform (or "skip") \cite{jiao2020tinybert, sun-etal-2019-patient} and Last \cite{mukherjee2021xtremedistiltransformers} strategies commonly used for KD where $L^T > L^S$:
\begin{equation}
g(i) = \{(i-1)\left\lceil\dfrac{L^T}{L^S}\right\rceil +1, L^T-L^S+i \} \\
\label{eq:mapping}
\end{equation}
We compare this method to common layer mapping strategies for HS distillation in \cref{app:hs-distil-perf}.

\paragraph{Neural Architecture Search} NAS automates the process of selecting the best candidate architecture for a given task. The \textbf{search space} defines the set of possible candidate architectures. For a transformer model, this may include components of the transformer architecture. The goal of NAS is to pick within the  the search space an architecture which performs optimally, using a \textbf{search strategy} which makes the process more efficient than a brute-force search. NAS typically employs algorithms such as Reinforcement Learning \cite{zoph2017neural, zoph2018learning, tan2019mnasnet}, Evolutionary Algorithms \cite{so2019evolved} or Differential Search \cite{liu2019darts} in order to optimize the search process over a large, multi-dimensional search space. Since training candidate architectures on the entire  target pipeline is computationally expensive, a common \textbf{performance evaluation} strategy is to conduct only a portion of the training on a "proxy set" \protect\cite{liu2019darts,zhou2020econas,na2021accelerating}. This constitutes training on a reduced version of the corpus for a limited number of epochs, and must be representative of the performance on the entire training pipeline.

\section{KD-NAS: KD Aware Neural Architecture Search}

We describe the design of our KD guided NAS process (KD-NAS), which uses NAS to find the optimal student architecture for task-agnostic distillation from a given teacher. \cref{fig-kdnas} shows the overall process.

\begin{table}
\centering
\begin{adjustbox}{max width=0.5\textwidth}
\begin{tabular}{cc}
\hline
 \textbf{Parameter}& \textbf{Candidate Values}\\
\hline
Hidden Layers & [3, 4, 6, 10, 12] \\
Attention Heads & [2, 3, 4, 6, 12] \\
Intermediate Size & [384, 512, 576, 768, 1024, 1536, 2048, 3072] \\
Hidden Activation & [gelu, relu, silu]\\
\hline
\end{tabular}
\end{adjustbox}
\caption{\label{search-space}
Search Space Candidates for KD-NAS, consisting of 2400 student architectures.}
\end{table}

\paragraph{Search Space}

We define a five-dimensional search space for the student transformer model, comprising of the number of \textit{hidden layers}, number of \textit{attention heads}, \textit{hidden size}, \textit{intermediate (FFN) size}, and hidden \textit{activation function}, as shown in \cref{search-space}. This space includes 2400 possible student architectures. Note, in this work we use the terms \textit{state} and \textit{architecture} interchangeably.

\paragraph{Performance Evaluation System}\label{perform-eval}
To estimate the reward, each candidate student model is distilled from the teacher on a proxy set. We define a 4-epoch \textit{mini-KD} process, using 30\% of the training corpus. This reduces the time to distil each student, thus being a practical method of evaluating multiple candidate architectures. We obtain this choice of proxy set by experimenting with different combinations of data size and training epochs, and select the smallest process that maintains the relative ranking of distillation loss for a set of randomly selected architectures from the search space.

\paragraph{Search Strategy}

This work implements a reinforcement learning based approach for NAS, with the following components: 

\subparagraph{A. Feedback Controller}\label{controller-design}
We train a Long Short Term Memory (LSTM) \protect\cite{LSTM, LSTMBakker} Controller model to predict the \emph{reward} given a state and a memory of performance of previously explored states. Specifically, in each NAS episode $ep_n$, we pass as input the current state being explored, and two additional states: the Global Best State (the highest-reward architecture from all prior episodes $ep_{i \in [1,n-1]}$), and the Previous Best State (the highest-reward architecture from episode $ep_{n-1}$). 

\subparagraph{B. Search Mechanism}
Each \emph{episode} of KD-NAS comprises of the following steps:
\begin{enumerate}[topsep=0pt, itemsep=0pt, leftmargin=.15in, parsep=0pt]
    \item Each episode generates $N$ candidate states, including those selected by the controller and those selected for random exploration. Initially, the controller benefits from more \emph{exploration} of the search space, which is controlled by the exploration ratio $\epsilon$\footnote{Thus, in each episode, $\epsilon \cdot N$ states chosen are random, and the controller model returns the remaining $(1-\epsilon) \cdot N$ states predicted to have the highest reward.}. We decay $\epsilon_1=1$ to $\epsilon_{min} = 0.05$ with a decay rate of $0.05$ per episode.
    \item The mini-KD process is performed for each candidate state $s$, to obtain the reward $reward(S_s)$ of the resulting distilled student $S_s$.
    \item The \emph{(state,reward)} pairs are concatenated with \emph{(GlobalBestState, PreviousBestState)} and is used as training data to minimize the Controller loss $L_C$, i.e, the MSE between the actual reward of the model after distillation on the proxy set, $reward(S_s)$, and the predicted reward $reward_{pred}(s)$ for all states $s$ belonging to the candidate states $CS$ of that episode:
    \begin{equation}
        L_C = \frac{1}{2} \sum_{s \in CS} \|reward(S_s) - reward_{pred}(s) \|^2
    \end{equation}
\end{enumerate}
This process is repeated for $M$ episodes, after which the explored architectures with the highest reward are selected.

\subparagraph{C. Reward Function}
The reward is a function of the student's latency  of inference on a CPU $lat(S_s)$, and the distillation loss $\mathcal{L}_{\mathrm{HS}}$ (\cref{eq:hs_transfer_1_to_n}) between the student $S_s$ and teacher $T$. The latency is normalized using the maximum target latency, i.e., a fraction of the teacher latency $lat(T)$.  The reward is similar to the optimization goal of \citet{tan2019mnasnet}:
\begin{equation}
reward(S_s) = (1-\mathcal{L}_{\mathrm{HS}}(S_s, T)) \cdot \left(\frac{lat(S_s)}{\beta * lat(T)}\right)^\alpha 
\label{eq:reward}
\end{equation}

The hyperparameters $\alpha$ and $\beta$ control the relative weight of latency in the optimization goal, and the maximum target latency respectively.

\section{Experiments and Results}

\begin{table}[t!]
\centering
\begin{adjustbox}{width=0.5\textwidth}
\begin{tabular}{c|c|c|cc}
\hline
 \textbf{Model}& \textbf{Architecture} & \textbf{Params}& \multicolumn{2}{c}{\textbf{Latency (ms)} $\downarrow$ }  \\
 & & \textbf{(M)} & CPU & GPU\\
\hline
KDNAS\textsubscript{Arch1} & 3,12,384,1024,gelu	& 100 & 5.97\textsubscript{\textpm0.04} &3.25\textsubscript{\textpm0.01}\\
KDNAS\textsubscript{Arch2} & 4,4,288,768,gelu	& 76 & 6.12\textsubscript{\textpm0.03} &3.49\textsubscript{\textpm0.02}\\
KDNAS\textsubscript{Arch3} & 4,12,576,768,gelu	& 153& 9.19\textsubscript{\textpm0.03}&	3.98\textsubscript{\textpm0.02}\\
\hline
6L-384H\textsuperscript{$\dagger$} & 6,12,384,1536,gelu	&107  & 12.41\textsubscript{\textpm0.03}&	5.99\textsubscript{\textpm0.07}\\
6L-768H\textsuperscript{$\dagger$} & 6,12,768,3072,gelu	& 234& 33.21\textsubscript{\textpm0.07} &6.01\textsubscript{\textpm0.06} \\
DistilBERT&6,12,768,3072,gelu	& 134\footnotemark& 30.93\textsubscript{\textpm0.03}	&5.64\textsubscript{\textpm0.04} \\
\hline
Teacher & 12,12,768,3072,gelu	& 277& 64.98\textsubscript{\textpm0.51}&	9.47\textsubscript{\textpm0.01}\\
\hline
\end{tabular}
\end{adjustbox}
\caption{\label{tab:nas-model}
Model Architectures, depicted as [hidden layers, attention heads, hidden size, intermediate size, activation function]. Latency measured on an A100 GPU and 4-core CPU, over 3 runs. $^\dagger$architectures distilled on our pipeline.}
\end{table}

\footnotetext{Multilingual DistilBERT from Huggingface has a smaller number of parameters that 6L-768H with the same architecture due its smaller vocabulary size (130K vs 250K).}

\paragraph{Baselines}
For a fair analysis of our technique, we compare our NAS-recommended architectures to two popular student architectures- a 6 layer, 768 hidden size (6L-768H) architecture \cite{sanh2020distilbert}, and a smaller 6 layer, 384 hidden size (6L-384H) architecture \cite{wang2020minilm, mukherjee2021xtremedistiltransformers, wang-etal-2021-minilmv2}, distilled on our own pipeline. We compare the \textit{architecture} of our KD-NAS models with common open source baselines, and demonstrate an optimal trade-off between speed and performance with our models.  Additionally, we compare our architectures and distillation method to the popular open source Multilingual Distil-BERT model\footnote{\Verb:distilbert-base-multilingual-cased: from Huggingface \cite{wolf2020huggingfaces}.}. These architectures, along with those recommended by KD-NAS, are described in Table \ref{tab:nas-model}.

\subsection{KD-NAS Implementation}
\paragraph{Data} We conduct a \emph{multilingual, task-agnostic} distillation to produce general purpose students, on a subset of 7M sentences from the multilingual CC100 dataset \cite{conneau2020unsupervised}\footnote{For training details, refer to \cref{app:hyperparameter}.}.

\paragraph{Architecture Search}

We conduct KD-NAS to find the best student architecture to distil from an in-house XLM-Roberta Base \cite{conneau2020unsupervised} teacher over $15$ episodes (each producing $20$ candidate architectures) using the proxy set \textit{mini-KD} as an estimation of distillation loss. KD-NAS is guided by the reward function \cref{eq:reward}, with $\alpha=-0.06$, $\beta=0.6$\footnote{We take a smaller value of $\alpha$ than proposed by \citep{tan2019mnasnet}, resulting in a reward that favors lower loss more.}. For consistent and comparable latency measurements, we create a lookup table of the latencies of all randomly-initialized students on a single CPU\footnote{We calculate the latency of a single inference, averaged over 10K sentences (batch size 1), across 3 runs.}.

We obtain the top $3$ architectures recommended\footnote{The controller predictions include some random exploration, however, the selected models have been recommended across multiple episodes, indicating that the controller is confident in the performance of these models, even if they may have been initially been candidates of exploration.} by the KD-NAS process, (KDNAS\textsubscript{Arch1}, KDNAS\textsubscript{Arch2}, KDNAS\textsubscript{Arch3}), with different latency budgets, as described in Table \ref{tab:nas-model}. To evaluate the efficacy of our NAS system,  we compare it to random sampling over the defined search space \cite{yang2020nas}. The average loss, latency, and reward after the distillation of 3 randomly selected student architectures over 3 seeds have been shown in Table \ref{tab:random-sampling}. As seen, KD-NAS models outperform the random sampling models in all metrics, encouraging the use of NAS over random sampling.

\begin{table}[t!]
\centering
\begin{adjustbox}{max width=0.5\textwidth}
\begin{tabular}{cccc}
\hline
 \multirow{2}{*}{\textbf{Model}}& \textbf{Distillation}  & \textbf{CPU}& \textbf{Reward} \\
 & \textbf{Loss}$\downarrow$ & \textbf{Latency (ms)}$\downarrow$ & $\uparrow$\\
\hline
Random\textsubscript{Seed1}	& 0.051\textsubscript{\textpm0.01} & 13.20\textsubscript{\textpm6.0} &1.02\textsubscript{\textpm0.03}\\
Random\textsubscript{Seed2}	& 0.050\textsubscript{\textpm0.01}& 11.40\textsubscript{\textpm3.6} & 1.03\textsubscript{\textpm0.02} \\
Random\textsubscript{Seed3}	& 0.057\textsubscript{\textpm0.01}& 14.17\textsubscript{\textpm3.1} &1.01\textsubscript{\textpm0.02} \\
\hline
KDNAS\textsubscript{Arch1} &0.050  &5.97 &1.07 \\
KDNAS\textsubscript{Arch2} &0.049 &6.12 &1.07 \\
KDNAS\textsubscript{Arch1} & 0.043& 9.19& 1.05\\
\hline
\end{tabular}
\end{adjustbox}
\caption{\label{tab:random-sampling}
Random Sampling vs KD-NAS. Avg. metrics shown for 3 random architectures per seed for 3 seeds.}
\end{table}

\begin{table*}[ht!]
\centering
\begin{adjustbox}{max width=\textwidth}
\begin{tabular}{c|c|cc|ccc|ccc|c}
\hline
\textbf{Model} & \textbf{Params} & \multicolumn{2}{c|}{\textbf{Latency} $\downarrow$}  & \multicolumn{3}{c|}{\textbf{Monolingual Performance} $\uparrow$}  & \multicolumn{3}{c|}{\textbf{Multilingual Performance} $\uparrow$}& {\textbf{Avg.} $\uparrow$} \\
 & \textbf{(M)}& CPU & GPU  & GLUE & SQuAD & CoNLL & XNLI & TyDi & NER\textsuperscript{$\star$} &   \\
\hline
KDNAS\textsubscript{Arch1} & 100 & 5.97  &3.25 & 76.1\textsubscript{\textpm0.6}&	55.5\textsubscript{\textpm0.3}&	90.0\textsubscript{\textpm0.1}&	57.9\textsubscript{\textpm0.5}&	56.3\textsubscript{\textpm0.4} & 71.3\textsubscript{\textpm0.1} & 67.9 \\
KDNAS\textsubscript{Arch2} & 76 & 6.12  & 3.49 & 76.1\textsubscript{\textpm0.3}&	56.9\textsubscript{\textpm0.2}&	90.0\textsubscript{\textpm0.3}&	58.3\textsubscript{\textpm0.6} & 58.7\textsubscript{\textpm0.3}&	72.0\textsubscript{\textpm0.4}&	68.6 \\
KDNAS\textsubscript{Arch3} & 153& 9.19 &	3.98 & 79.6\textsubscript{\textpm0.6}&	62.7\textsubscript{\textpm0.1}&	92.1\textsubscript{\textpm0.2}&	62.7\textsubscript{\textpm0.5}&	66.8\textsubscript{\textpm0.6} & 72.8\textsubscript{\textpm0.1}& 72.8 \\
\hline
6L-384H\textsuperscript{$\dagger$}&107  & 12.41 &	5.99 & 78.7\textsubscript{\textpm0.8}&	62.4\textsubscript{\textpm0.1}&	91.4\textsubscript{\textpm0.2}&	62.7\textsubscript{\textpm0.4}&	63.4\textsubscript{\textpm0.5}& 73.1\textsubscript{\textpm0.1} & 71.9\\
6L-768H\textsuperscript{$\dagger$}&234  & 33.21 &	6.01  & 83.1\textsubscript{\textpm0.3}&	68.0\textsubscript{\textpm0.3}&	93.9\textsubscript{\textpm0.1}&	67.0\textsubscript{\textpm0.4}&	72.7\textsubscript{\textpm0.6}&	74.2\textsubscript{\textpm0.1}& 76.5 \\
DistilBERT& 134& 30.93	&5.64 & 81.4\textsubscript{\textpm0.5}&	68.9 \textsubscript{\textpm0.4}&	94.2\textsubscript{\textpm0.1}&	59.4\textsubscript{\textpm0.5}&	69.8\textsubscript{\textpm0.2}& 74.2\textsubscript{\textpm0.1}& 74.7\\
\hline
Random\textsubscript{Seed1}	 & 115 & 13.20 & 5.31 & 62.6\textsubscript{\textpm6.2} &	43.6\textsubscript{\textpm0.9} &	86.3\textsubscript{\textpm7.6}&	56.2\textsubscript{\textpm5.7}& 49.3\textsubscript{\textpm13.14}& 70.9\textsubscript{\textpm3.6}& 61.5  \\
Random\textsubscript{Seed2}	 & 123 & 11.40 & 5.89 & 76.1\textsubscript{\textpm3.8} & 56.1\textsubscript{\textpm9.2}&	89.1\textsubscript{\textpm1.0}&	60.6\textsubscript{\textpm5.3}&61.0\textsubscript{\textpm4.0} &71.6\textsubscript{\textpm0.7} &69.1 \\
Random\textsubscript{Seed3}	 & 91 & 14.17 & 6.38& 66.5\textsubscript{\textpm3.0} &	44.5\textsubscript{\textpm2.0}& 88.6\textsubscript{\textpm3.1}&	58.0\textsubscript{\textpm1.8}&53.9\textsubscript{\textpm5.3} &72.5\textsubscript{\textpm1.4} & 64.1\\
\hline
Teacher & 277& 64.98 &	9.47 & 84.8\textsubscript{\textpm0.3}&	77.3\textsubscript{\textpm0.5}&	94.2\textsubscript{\textpm0.1}&	70.9\textsubscript{\textpm0.8}&	78.1\textsubscript{\textpm0.7}& 76.1\textsubscript{\textpm0.1} &80.2\\
\hline
\end{tabular}
\end{adjustbox}
\caption{\label{tab:final-compare}
KD-NAS performance, compared to baselines and random sampling. We report avg. GLUE score across all tasks, avg. XNLI score across all languages, and F1 score of SQuAD, TyDi and CoNLL. Latency measured in ms for an A100 GPU and 4-core CPU. Results shown as avg. across 3 runs. ($^\dagger$baselines distilled on our pipeline,  $^\star$internal dataset)}
\end{table*}

\subsection{Results}
After obtaining the KD-NAS recommendations, we distil the models over the entire training pipeline, with the objective described in \cref{eq:hs_transfer_1_to_n}, to obtain general purpose models. We finetune these models on both monolingual (English) and multilingual benchmarks, and on tasks including text classification (GLUE\footnote{We report the average performance on all tasks of GLUE except CoLA. Distilled models perform poorly on CoLA- this may be because CoLA is a \textit{syntactic} task, as opposed to the other semantic tasks in the benchmark. For results on all tasks, please refer to \cref{app:results}.} \cite{wang2019glue}, XNLI \cite{conneau2018xnli}), Question Answering (SQuAD \cite{rajpurkar2016squad}, TyDiQA \cite{clark2020tydi}), and Named Entity Recognition (CoNLL \cite{tjong2003conll}). We also evaluate on an internal dataset for multilingual Named Entity Recognition, which consists of $17$ entities across $24$ languages (a detailed description of this dataset is given in \cref{app:wnlpner}).

We report our results in \cref{tab:final-compare}, which shows our KD-NAS- recommended models have the least latency of inference on both CPU and GPU, while maintaining good performance. Specifically,
\begin{itemize}[topsep=0pt, itemsep=0pt, leftmargin=.2in, parsep=0pt]
    \item Our best performing model, KDNAS\textsubscript{Arch3}, obtains equivalent or better performance to our 6L-384H baseline on all tasks, while being significantly faster.
    \item Compared to the 6L-768H architectures, our models show a slight degradation in performance, however, they show a significant gain in speed. Moreover, the average performance of random sampling from the search space is always sub-optimal compared to the KD-NAS models. 
    Thus, KD-NAS can automatically find an optimal trade-off between effectiveness and efficiency, which is expensive to do manually.
    \item Our proposed multi-layer HS distillation method outperforms multilingual DistilBERT, with the most significant gains seen for multilingual tasks, demonstrating the advantage of transferring the knowledge from multiple teacher layers to each student layer. Notably, this method achieves good results without the need of pre-training (which is computationally expensive) or initializing layers from the teacher (which constrains the architecture).
\end{itemize}

\section{Applications}

$KDNAS_{Arch3}$, which we refer to as the \emph{Piccolo} Architecture, is significantly faster than the teacher, while maintaining good performance. After further experiments, we find that distilling the self-attention relations as proposed in MiniLM-v2 \cite{wang-etal-2021-minilmv2} improves performance \cite{udagawa2023comparative}\footnote{Performance described in \cref{app:minilmv2-perf}. We aim to conduct KD-NAS with a MiniLM-v2 based objective as future work.}. We use this distillation objective to produce a model that has been deployed for natural language understanding applications requiring low latency and high throughput, namely:
\begin{itemize}[topsep=0pt, itemsep=0pt, leftmargin=.2in, parsep=0pt]
    \item Low Latency NLP: after being finetuned on downstream tasks (e.g. entity extraction, sentiment analysis, relation detection) the Piccolo model provides a 7.6x speedup on CPU (2.5x on A100 GPU) relative to the 12-layer teacher, while retaining 93\%  performance on average. It also results in a 5.9x speedup in finetuning time on average. This allows easy deployment on CPU, edge devices, or low-latency GPU applications, and has been deployed as part of 3 software offerings \cite{Lang_2023, Heidloff_2023}.
    \item Hate/Abuse/Profanity (HAP) filtering for data pre-processing: Language models are trained on data originating from internet sources, which may contain profane language. Using uncivil content for training may lead to the model producing similar hateful content as well, which is not suitable for business use. The Piccolo Model, finetuned to identify hate-speech \cite{poletto21hatespeech}, is used to efficiently filter a large corpus of training data for HAP. Using this model results in faster inference (7x on CPU, 2x on GPU), larger throughput (3.5x) and a lower memory consumption (4x fewer parameters) than a 12-layer model, while maintaining equivalent performance, and has been deployed in internal and external software offerings.
\end{itemize}

\section{Related Work}

Our work is most comparable to prior work that uses KD to guide NAS for language models. AdaBERT \cite{chen2021adabert} compresses monolingual BERT models into models using differentiable NAS that also combines a distillation loss into the search objective. However, they conduct their search in a \textit{task-specific} manner, where the KD loss is determined by using probes to hierarchically decompose the task-useful knowledge from the teacher model, and then distilling it. While this may be able to achieve a higher compression rate, is not effective for producing general purpose students that can be further finetuned on task-specific data, which are more useful in industrial settings. NAS-BERT \cite{Xu_2021_nasbert} produces task-agnostic, compressed monolingual BERT models for different latency budgets, by training a large supernet using block-wise training. While each block of the supernet is trained using KD to mimic the corresponding teacher block, we use KD explicitly to \emph{guide} the search process and formulate the NAS reward, and conduct our search on the entire student architecture. Moreover, our Feedback Controller model determines the next states to be explored by predicting the rewards of the given state along with information regarding well-performing states seen in the past, which is a contrast from the probabilistic actor-critic agent proposed in previous work \cite{liu2020search, bashivan2019teacher, Wang_2021}. While most prior art focuses on monolingual (English) models, we expand search to the multilingual domain in a task-agnostic setting.

\section{Conclusion}

We develop KD-NAS, a method to non-trivially adopt NAS to optimize for the KD objective, thus transferring task-agnostic knowledge from a large model into an optimal architecture within a large search space. After distillation, our KD-NAS models show significant speedup over a hand-crafted baseline while maintaining performance. Larger students distilled with our task-agnostic multi-layer hidden state transfer objective outperforms an open-source model with the same architecture. We illustrate how a high performing KD-NAS model has been deployed for low-latency applications.

\begin{center}
\bibliography{emnlp2023}

\begin{thebibliography}{55}
\expandafter\ifx\csname natexlab\endcsname\relax\def\natexlab#1{#1}\fi

\bibitem[{Ba and Caruana(2014)}]{ba_2014}
Jimmy Ba and Rich Caruana. 2014.
\newblock \href
  {https://proceedings.neurips.cc/paper/2014/file/ea8fcd92d59581717e06eb187f10666d-Paper.pdf}
  {Do deep nets really need to be deep?}
\newblock In \emph{Advances in Neural Information Processing Systems},
  volume~27. Curran Associates, Inc.

\bibitem[{Bakker(2002)}]{LSTMBakker}
Bram Bakker. 2002.
\newblock \href
  {https://proceedings.neurips.cc/paper/2001/file/a38b16173474ba8b1a95bcbc30d3b8a5-Paper.pdf}
  {Reinforcement learning with long short-term memory}.
\newblock In \emph{Advances in Neural Information Processing Systems},
  volume~14. MIT Press.

\bibitem[{Bashivan et~al.(2019)Bashivan, Tensen, and
  DiCarlo}]{bashivan2019teacher}
Pouya Bashivan, Mark Tensen, and James~J DiCarlo. 2019.
\newblock \href {http://arxiv.org/abs/1808.01405} {Teacher guided architecture
  search}.

\bibitem[{Brown et~al.(2020)Brown, Mann, Ryder, Subbiah, Kaplan, Dhariwal,
  Neelakantan, Shyam, Sastry, Askell et~al.}]{brown2020language}
Tom Brown, Benjamin Mann, Nick Ryder, Melanie Subbiah, Jared~D Kaplan, Prafulla
  Dhariwal, Arvind Neelakantan, Pranav Shyam, Girish Sastry, Amanda Askell,
  et~al. 2020.
\newblock Language models are few-shot learners.
\newblock \emph{Advances in neural information processing systems},
  33:1877--1901.

\bibitem[{Chen et~al.(2021)Chen, Li, Qiu, Wang, Li, Ding, Deng, Huang, Lin, and
  Zhou}]{chen2021adabert}
Daoyuan Chen, Yaliang Li, Minghui Qiu, Zhen Wang, Bofang Li, Bolin Ding, Hongbo
  Deng, Jun Huang, Wei Lin, and Jingren Zhou. 2021.
\newblock \href {http://arxiv.org/abs/2001.04246} {Adabert: Task-adaptive bert
  compression with differentiable neural architecture search}.

\bibitem[{Clark et~al.(2020{\natexlab{a}})Clark, Choi, Collins, Garrette,
  Kwiatkowski, Nikolaev, and Palomaki}]{clark2020tydi}
Jonathan~H. Clark, Eunsol Choi, Michael Collins, Dan Garrette, Tom Kwiatkowski,
  Vitaly Nikolaev, and Jennimaria Palomaki. 2020{\natexlab{a}}.
\newblock \href {http://arxiv.org/abs/2003.05002} {Tydi qa: A benchmark for
  information-seeking question answering in typologically diverse languages}.

\bibitem[{Clark et~al.(2020{\natexlab{b}})Clark, Luong, Le, and
  Manning}]{clark2020electra}
Kevin Clark, Minh-Thang Luong, Quoc~V. Le, and Christopher~D. Manning.
  2020{\natexlab{b}}.
\newblock \href {http://arxiv.org/abs/2003.10555} {Electra: Pre-training text
  encoders as discriminators rather than generators}.

\bibitem[{Conneau et~al.(2020)Conneau, Khandelwal, Goyal, Chaudhary, Wenzek,
  Guzmán, Grave, Ott, Zettlemoyer, and Stoyanov}]{conneau2020unsupervised}
Alexis Conneau, Kartikay Khandelwal, Naman Goyal, Vishrav Chaudhary, Guillaume
  Wenzek, Francisco Guzmán, Edouard Grave, Myle Ott, Luke Zettlemoyer, and
  Veselin Stoyanov. 2020.
\newblock \href {http://arxiv.org/abs/1911.02116} {Unsupervised cross-lingual
  representation learning at scale}.

\bibitem[{Conneau et~al.(2018)Conneau, Lample, Rinott, Williams, Bowman,
  Schwenk, and Stoyanov}]{conneau2018xnli}
Alexis Conneau, Guillaume Lample, Ruty Rinott, Adina Williams, Samuel~R.
  Bowman, Holger Schwenk, and Veselin Stoyanov. 2018.
\newblock \href {http://arxiv.org/abs/1809.05053} {Xnli: Evaluating
  cross-lingual sentence representations}.

\bibitem[{Devlin et~al.(2019)Devlin, Chang, Lee, and
  Toutanova}]{devlin2019bert}
Jacob Devlin, Ming-Wei Chang, Kenton Lee, and Kristina Toutanova. 2019.
\newblock \href {http://arxiv.org/abs/1810.04805} {Bert: Pre-training of deep
  bidirectional transformers for language understanding}.

\bibitem[{Elsken et~al.(2019)Elsken, Metzen, and Hutter}]{elsken2019neural}
Thomas Elsken, Jan~Hendrik Metzen, and Frank Hutter. 2019.
\newblock \href {http://arxiv.org/abs/1808.05377} {Neural architecture search:
  A survey}.

\bibitem[{Fedus et~al.(2021)Fedus, Zoph, and Shazeer}]{fedus2021switch}
William Fedus, Barret Zoph, and Noam Shazeer. 2021.
\newblock \href {http://arxiv.org/abs/2101.03961} {Switch transformers: Scaling
  to trillion parameter models with simple and efficient sparsity}.

\bibitem[{Heidloff(2023)}]{Heidloff_2023}
Niklas Heidloff. 2023.
\newblock \href
  {https://heidloff.net/article/ibm-announces-new-foundation-model-capabilities/}
  {Ibm announces new foundation model capabilities}.

\bibitem[{Hinton et~al.(2015)Hinton, Vinyals, and Dean}]{hinton2015distilling}
Geoffrey Hinton, Oriol Vinyals, and Jeff Dean. 2015.
\newblock \href {http://arxiv.org/abs/1503.02531} {Distilling the knowledge in
  a neural network}.

\bibitem[{Hochreiter and Schmidhuber(1997)}]{LSTM}
Sepp Hochreiter and Jürgen Schmidhuber. 1997.
\newblock \href {https://doi.org/10.1162/neco.1997.9.8.1735} {{Long Short-Term
  Memory}}.
\newblock \emph{Neural Computation}, 9(8):1735--1780.

\bibitem[{Huang et~al.(2015)Huang, Xu, and Yu}]{huang2015bidirectional}
Zhiheng Huang, Wei Xu, and Kai Yu. 2015.
\newblock \href {http://arxiv.org/abs/1508.01991} {Bidirectional lstm-crf
  models for sequence tagging}.

\bibitem[{Jafari et~al.(2021)Jafari, Rezagholizadeh, Sharma, and
  Ghodsi}]{jafari-etal-2021-annealing}
Aref Jafari, Mehdi Rezagholizadeh, Pranav Sharma, and Ali Ghodsi. 2021.
\newblock \href {https://doi.org/10.18653/v1/2021.eacl-main.212} {Annealing
  knowledge distillation}.
\newblock In \emph{Proceedings of the 16th Conference of the European Chapter
  of the Association for Computational Linguistics: Main Volume}, pages
  2493--2504, Online. Association for Computational Linguistics.

\bibitem[{Jiao et~al.(2021)Jiao, Chang, Yin, Shang, Jiang, Chen, Li, Wang, and
  Liu}]{jiao2021improving}
Xiaoqi Jiao, Huating Chang, Yichun Yin, Lifeng Shang, Xin Jiang, Xiao Chen,
  Linlin Li, Fang Wang, and Qun Liu. 2021.
\newblock Improving task-agnostic bert distillation with layer mapping search.
\newblock \emph{Neurocomputing}, 461:194--203.

\bibitem[{Jiao et~al.(2020)Jiao, Yin, Shang, Jiang, Chen, Li, Wang, and
  Liu}]{jiao2020tinybert}
Xiaoqi Jiao, Yichun Yin, Lifeng Shang, Xin Jiang, Xiao Chen, Linlin Li, Fang
  Wang, and Qun Liu. 2020.
\newblock \href {http://arxiv.org/abs/1909.10351} {Tinybert: Distilling bert
  for natural language understanding}.

\bibitem[{Kaplan et~al.(2020)Kaplan, McCandlish, Henighan, Brown, Chess, Child,
  Gray, Radford, Wu, and Amodei}]{kaplan2020scaling}
Jared Kaplan, Sam McCandlish, Tom Henighan, Tom~B Brown, Benjamin Chess, Rewon
  Child, Scott Gray, Alec Radford, Jeffrey Wu, and Dario Amodei. 2020.
\newblock Scaling laws for neural language models.
\newblock \emph{arXiv preprint arXiv:2001.08361}.

\bibitem[{Ko et~al.(2023)Ko, Park, Jeong, Hong, Ahn, Chang, and
  Yun}]{ko2023revisiting}
Jongwoo Ko, Seungjoon Park, Minchan Jeong, Sukjin Hong, Euijai Ahn, Du-Seong
  Chang, and Se-Young Yun. 2023.
\newblock \href {http://arxiv.org/abs/2302.01530} {Revisiting intermediate
  layer distillation for compressing language models: An overfitting
  perspective}.

\bibitem[{Lang(2023)}]{Lang_2023}
Alexander Lang. 2023.
\newblock \href
  {https://medium.com/@alex.lang/fair-is-fast-and-fast-is-fair-ibm-slate-foundation-models-for-nlp-3508412a4b04}
  {Fair is fast, and fast is fair: Ibm slate foundation models for nlp}.

\bibitem[{Li et~al.(2020)Li, Liu, Zhao, Xu, Yang, and Jin}]{li-etal-2020-bert}
Jianquan Li, Xiaokang Liu, Honghong Zhao, Ruifeng Xu, Min Yang, and Yaohong
  Jin. 2020.
\newblock \href {https://doi.org/10.18653/v1/2020.emnlp-main.242}
  {{BERT}-{EMD}: Many-to-many layer mapping for {BERT} compression with earth
  mover{'}s distance}.
\newblock In \emph{Proceedings of the 2020 Conference on Empirical Methods in
  Natural Language Processing (EMNLP)}, pages 3009--3018, Online. Association
  for Computational Linguistics.

\bibitem[{Liu et~al.(2019{\natexlab{a}})Liu, Simonyan, and Yang}]{liu2019darts}
Hanxiao Liu, Karen Simonyan, and Yiming Yang. 2019{\natexlab{a}}.
\newblock \href {http://arxiv.org/abs/1806.09055} {Darts: Differentiable
  architecture search}.

\bibitem[{Liu et~al.(2019{\natexlab{b}})Liu, Ott, Goyal, Du, Joshi, Chen, Levy,
  Lewis, Zettlemoyer, and Stoyanov}]{liu2019roberta}
Yinhan Liu, Myle Ott, Naman Goyal, Jingfei Du, Mandar Joshi, Danqi Chen, Omer
  Levy, Mike Lewis, Luke Zettlemoyer, and Veselin Stoyanov. 2019{\natexlab{b}}.
\newblock \href {http://arxiv.org/abs/1907.11692} {Roberta: A robustly
  optimized bert pretraining approach}.

\bibitem[{Liu et~al.(2020)Liu, Jia, Tan, Vemulapalli, Zhu, Green, and
  Wang}]{liu2020search}
Yu~Liu, Xuhui Jia, Mingxing Tan, Raviteja Vemulapalli, Yukun Zhu, Bradley
  Green, and Xiaogang Wang. 2020.
\newblock \href {http://arxiv.org/abs/1911.09074} {Search to distill: Pearls
  are everywhere but not the eyes}.

\bibitem[{Loshchilov and Hutter(2019)}]{loshchilov2019adamw}
Ilya Loshchilov and Frank Hutter. 2019.
\newblock \href {http://arxiv.org/abs/1711.05101} {Decoupled weight decay
  regularization}.

\bibitem[{Mukherjee et~al.(2021)Mukherjee, Awadallah, and
  Gao}]{mukherjee2021xtremedistiltransformers}
Subhabrata Mukherjee, Ahmed~Hassan Awadallah, and Jianfeng Gao. 2021.
\newblock \href {http://arxiv.org/abs/2106.04563} {Xtremedistiltransformers:
  Task transfer for task-agnostic distillation}.

\bibitem[{Na et~al.(2021)Na, Mok, Choe, and Yoon}]{na2021accelerating}
Byunggook Na, Jisoo Mok, Hyeokjun Choe, and Sungroh Yoon. 2021.
\newblock \href {http://arxiv.org/abs/2106.04784} {Accelerating neural
  architecture search via proxy data}.

\bibitem[{Poletto et~al.(2021)Poletto, Basile, Sanguinetti, Bosco, and
  Patti}]{poletto21hatespeech}
Fabio Poletto, Valerio Basile, Manuela Sanguinetti, Cristina Bosco, and Viviana
  Patti. 2021.
\newblock \href {https://doi.org/10.1007/s10579-020-09502-8} {Resources and
  benchmark corpora for hate speech detection: a systematic review}.
\newblock \emph{Language Resources and Evaluation}, 55(2):477--523.

\bibitem[{Rajpurkar et~al.(2016)Rajpurkar, Zhang, Lopyrev, and
  Liang}]{rajpurkar2016squad}
Pranav Rajpurkar, Jian Zhang, Konstantin Lopyrev, and Percy Liang. 2016.
\newblock \href {http://arxiv.org/abs/1606.05250} {Squad: 100,000+ questions
  for machine comprehension of text}.

\bibitem[{Sanh et~al.(2020)Sanh, Debut, Chaumond, and
  Wolf}]{sanh2020distilbert}
Victor Sanh, Lysandre Debut, Julien Chaumond, and Thomas Wolf. 2020.
\newblock \href {http://arxiv.org/abs/1910.01108} {Distilbert, a distilled
  version of bert: smaller, faster, cheaper and lighter}.

\bibitem[{So et~al.(2019)So, Liang, and Le}]{so2019evolved}
David~R. So, Chen Liang, and Quoc~V. Le. 2019.
\newblock \href {http://arxiv.org/abs/1901.11117} {The evolved transformer}.

\bibitem[{Sun et~al.(2019)Sun, Cheng, Gan, and Liu}]{sun-etal-2019-patient}
Siqi Sun, Yu~Cheng, Zhe Gan, and Jingjing Liu. 2019.
\newblock \href {https://doi.org/10.18653/v1/D19-1441} {Patient knowledge
  distillation for {BERT} model compression}.
\newblock In \emph{Proceedings of the 2019 Conference on Empirical Methods in
  Natural Language Processing and the 9th International Joint Conference on
  Natural Language Processing (EMNLP-IJCNLP)}, pages 4323--4332, Hong Kong,
  China. Association for Computational Linguistics.

\bibitem[{Tan et~al.(2019)Tan, Chen, Pang, Vasudevan, Sandler, Howard, and
  Le}]{tan2019mnasnet}
Mingxing Tan, Bo~Chen, Ruoming Pang, Vijay Vasudevan, Mark Sandler, Andrew
  Howard, and Quoc~V. Le. 2019.
\newblock \href {http://arxiv.org/abs/1807.11626} {Mnasnet: Platform-aware
  neural architecture search for mobile}.

\bibitem[{Tenney et~al.(2019)Tenney, Das, and Pavlick}]{tenney2019bert}
Ian Tenney, Dipanjan Das, and Ellie Pavlick. 2019.
\newblock \href {http://arxiv.org/abs/1905.05950} {Bert rediscovers the
  classical nlp pipeline}.

\bibitem[{Tijmen and Hinton(2012)}]{HintonRMSProp}
Tieleman Tijmen and Geoffrey Hinton. 2012.
\newblock \href
  {https://www.cs.toronto.edu/~tijmen/csc321/slides/lecture_slides_lec6.pdf}
  {Lecture 6.5-rmsprop: Divide the gradient by a running average of its recent
  magnitude.}

\bibitem[{Tjong Kim~Sang and De~Meulder(2003)}]{tjong2003conll}
Erik~F. Tjong Kim~Sang and Fien De~Meulder. 2003.
\newblock \href {https://aclanthology.org/W03-0419} {Introduction to the
  {C}o{NLL}-2003 shared task: Language-independent named entity recognition}.
\newblock In \emph{Proceedings of the Seventh Conference on Natural Language
  Learning at {HLT}-{NAACL} 2003}, pages 142--147.

\bibitem[{Turc et~al.(2019)Turc, Chang, Lee, and Toutanova}]{turc2019pretrain}
Iulia Turc, Ming-Wei Chang, Kenton Lee, and Kristina Toutanova. 2019.
\newblock \href {https://doi.org/10.48550/ARXIV.1908.08962} {Well-read students
  learn better: On the importance of pre-training compact models}.

\bibitem[{Udagawa et~al.(2023)Udagawa, Trivedi, Merler, and
  Bhattacharjee}]{udagawa2023comparative}
Takuma Udagawa, Aashka Trivedi, Michele Merler, and Bishwaranjan Bhattacharjee.
  2023.
\newblock A comparative analysis of task-agnostic distillation methods for
  compressing transformer language models.
\newblock In \emph{Proceedings of the 2023 Conference on Empirical Methods in
  Natural Language Processing: Industry Track}.

\bibitem[{Vaswani et~al.(2017)Vaswani, Shazeer, Parmar, Uszkoreit, Jones,
  Gomez, Kaiser, and Polosukhin}]{vaswani2017attention}
Ashish Vaswani, Noam Shazeer, Niki Parmar, Jakob Uszkoreit, Llion Jones,
  Aidan~N. Gomez, Lukasz Kaiser, and Illia Polosukhin. 2017.
\newblock \href {http://arxiv.org/abs/1706.03762} {Attention is all you need}.

\bibitem[{Wang et~al.(2019)Wang, Singh, Michael, Hill, Levy, and
  Bowman}]{wang2019glue}
Alex Wang, Amanpreet Singh, Julian Michael, Felix Hill, Omer Levy, and
  Samuel~R. Bowman. 2019.
\newblock \href {http://arxiv.org/abs/1804.07461} {Glue: A multi-task benchmark
  and analysis platform for natural language understanding}.

\bibitem[{Wang et~al.(2021)Wang, Bao, Huang, Dong, and
  Wei}]{wang-etal-2021-minilmv2}
Wenhui Wang, Hangbo Bao, Shaohan Huang, Li~Dong, and Furu Wei. 2021.
\newblock \href {https://doi.org/10.18653/v1/2021.findings-acl.188}
  {{M}ini{LM}v2: Multi-head self-attention relation distillation for
  compressing pretrained transformers}.
\newblock In \emph{Findings of the Association for Computational Linguistics:
  ACL-IJCNLP 2021}, pages 2140--2151, Online. Association for Computational
  Linguistics.

\bibitem[{Wang et~al.(2020)Wang, Wei, Dong, Bao, Yang, and
  Zhou}]{wang2020minilm}
Wenhui Wang, Furu Wei, Li~Dong, Hangbo Bao, Nan Yang, and Ming Zhou. 2020.
\newblock \href {http://arxiv.org/abs/2002.10957} {Minilm: Deep self-attention
  distillation for task-agnostic compression of pre-trained transformers}.

\bibitem[{Wang(2021)}]{Wang_2021}
Xiaobo Wang. 2021.
\newblock \href {https://ojs.aaai.org/index.php/AAAI/article/view/16387}
  {Teacher guided neural architecture search for face recognition}.
\newblock \emph{Proceedings of the AAAI Conference on Artificial Intelligence},
  35(4):2817--2825.

\bibitem[{Wang et~al.(2023)Wang, Weissweiler, Sch{\"u}tze, and
  Plank}]{wang-etal-2023-distill}
Xinpeng Wang, Leonie Weissweiler, Hinrich Sch{\"u}tze, and Barbara Plank. 2023.
\newblock \href {https://aclanthology.org/2023.acl-short.157} {How to distill
  your {BERT}: An empirical study on the impact of weight initialisation and
  distillation objectives}.
\newblock In \emph{Proceedings of the 61st Annual Meeting of the Association
  for Computational Linguistics (Volume 2: Short Papers)}, pages 1843--1852,
  Toronto, Canada. Association for Computational Linguistics.

\bibitem[{Wolf et~al.(2020)Wolf, Debut, Sanh, Chaumond, Delangue, Moi, Cistac,
  Rault, Louf, Funtowicz, Davison, Shleifer, von Platen, Ma, Jernite, Plu, Xu,
  Scao, Gugger, Drame, Lhoest, and Rush}]{wolf2020huggingfaces}
Thomas Wolf, Lysandre Debut, Victor Sanh, Julien Chaumond, Clement Delangue,
  Anthony Moi, Pierric Cistac, Tim Rault, Rémi Louf, Morgan Funtowicz, Joe
  Davison, Sam Shleifer, Patrick von Platen, Clara Ma, Yacine Jernite, Julien
  Plu, Canwen Xu, Teven~Le Scao, Sylvain Gugger, Mariama Drame, Quentin Lhoest,
  and Alexander~M. Rush. 2020.
\newblock \href {http://arxiv.org/abs/1910.03771} {Huggingface's transformers:
  State-of-the-art natural language processing}.

\bibitem[{Wu et~al.(2020)Wu, Passban, Rezagholizadeh, and
  Liu}]{wu-etal-2020-skip}
Yimeng Wu, Peyman Passban, Mehdi Rezagholizadeh, and Qun Liu. 2020.
\newblock \href {https://doi.org/10.18653/v1/2020.emnlp-main.74} {Why skip if
  you can combine: A simple knowledge distillation technique for intermediate
  layers}.
\newblock In \emph{Proceedings of the 2020 Conference on Empirical Methods in
  Natural Language Processing (EMNLP)}, pages 1016--1021, Online. Association
  for Computational Linguistics.

\bibitem[{Wu et~al.(2021)Wu, Rezagholizadeh, Ghaddar, Haidar, and
  Ghodsi}]{wu-etal-2021-universal}
Yimeng Wu, Mehdi Rezagholizadeh, Abbas Ghaddar, Md~Akmal Haidar, and Ali
  Ghodsi. 2021.
\newblock \href {https://doi.org/10.18653/v1/2021.emnlp-main.603}
  {Universal-{KD}: Attention-based output-grounded intermediate layer knowledge
  distillation}.
\newblock In \emph{Proceedings of the 2021 Conference on Empirical Methods in
  Natural Language Processing}, pages 7649--7661, Online and Punta Cana,
  Dominican Republic. Association for Computational Linguistics.

\bibitem[{Xu et~al.(2021)Xu, Tan, Luo, Song, Li, Qin, and
  Liu}]{Xu_2021_nasbert}
Jin Xu, Xu~Tan, Renqian Luo, Kaitao Song, Jian Li, Tao Qin, and Tie-Yan Liu.
  2021.
\newblock \href {https://doi.org/10.1145/3447548.3467262} {Nas-bert:
  Task-agnostic and adaptive-size bert compression with neural architecture
  search}.
\newblock In \emph{Proceedings of the 27th {ACM} {SIGKDD} Conference on
  Knowledge Discovery Data Mining}. {ACM}.

\bibitem[{Yang et~al.(2020)Yang, Esperan{\c{c}}a, and Carlucci}]{yang2020nas}
Antoine Yang, Pedro~M. Esperan{\c{c}}a, and Fabio~Maria Carlucci. 2020.
\newblock {NAS} evaluation is frustratingly hard.
\newblock In \emph{International Conference on Learning Representations
  (ICLR)}.

\bibitem[{Zagoruyko and Komodakis(2017)}]{zagoruyko2017paying}
Sergey Zagoruyko and Nikos Komodakis. 2017.
\newblock \href {http://arxiv.org/abs/1612.03928} {Paying more attention to
  attention: Improving the performance of convolutional neural networks via
  attention transfer}.

\bibitem[{Zhou et~al.(2020)Zhou, Zhou, Zhang, Loy, Yi, Zhang, and
  Ouyang}]{zhou2020econas}
Dongzhan Zhou, Xinchi Zhou, Wenwei Zhang, Chen~Change Loy, Shuai Yi, Xuesen
  Zhang, and Wanli Ouyang. 2020.
\newblock \href {http://arxiv.org/abs/2001.01233} {Econas: Finding proxies for
  economical neural architecture search}.

\bibitem[{Zoph and Le(2017)}]{zoph2017neural}
Barret Zoph and Quoc~V. Le. 2017.
\newblock \href {http://arxiv.org/abs/1611.01578} {Neural architecture search
  with reinforcement learning}.

\bibitem[{Zoph et~al.(2018)Zoph, Vasudevan, Shlens, and Le}]{zoph2018learning}
Barret Zoph, Vijay Vasudevan, Jonathon Shlens, and Quoc~V. Le. 2018.
\newblock \href {http://arxiv.org/abs/1707.07012} {Learning transferable
  architectures for scalable image recognition}.

\end{thebibliography}
\bibliographystyle{acl_natbib}
\end{center}

\appendix

\section{Hidden State Distillation} \label{app:hs-distil-perf}
This work presents a \textit{multi-layer} hidden state distillation approach, where each student layer learns from two teacher layers. While transferring hidden states, the choice of layer mapping is key, and is often a controversial choice \cite{wu-etal-2020-skip,jiao2021improving,ko2023revisiting}. Here, we compare \textit{single-layer mapping} strategies (where a student layer is mapped to only one teacher layer) with our \emph{Uniform+Last} multi-layer mapping strategy when distilling the 6L-768H student baseline. The single-layer mapping strategies studied are
\begin{enumerate}[topsep=0pt, itemsep=0pt, leftmargin=.2in, parsep=0pt]
    \item \emph{Last-1}: we distil the last teacher layer into the last student layer.
    \item \emph{Last}: we map each student layer $i$ to the last $i^{th}$ teacher layer \cite{mukherjee2021xtremedistiltransformers}.
    \item \emph{Uniform}: we choose every $k^{th}$ teacher layer, where $k = \lceil L^T/L^S \rceil$ (also known as the "skip" strategy) \cite{sanh2020distilbert, sun-etal-2019-patient}.
\end{enumerate}
We evaluate the distilled students on the GLUE benchmark \cite{wang2019glue} and the XNLI task \cite{conneau2018xnli}. The results in \cref{tab:hs_distillation} show the proposed Uniform+Last strategy outperforming all single layer mapping strategies, as it allows for the transfer of diverse knowledge from the teacher.

\section{Training Details and Hyperparameters}\label{app:hyperparameter}

\begin{table}[t!]
\centering
\begin{adjustbox}{max width=0.5\textwidth}
\begin{tabular}{c|c|c|c}
\hline
\textbf{Dataset} & \textbf{bsz}  & \textbf{lr} & \textbf{epochs}\\
 \hline
GLUE (-CoLA) & 32 & 2e-5 & 3 \\
CoLA & 32 & 2e-5 & 10 \\
SQuAD & 32 & 3e-5 & 6 \\
CoNLL & 32 & 5e-5 & 3 \\
XNLI & 32 & 2e-5& 5 \\
TyDi & 32 & 1e-5 & 10 \\
NER\textsuperscript{$\star$} & 32 & 5e-5 & 5 \\ 
\hline
\end{tabular}
\end{adjustbox}
\caption{\label{tab:hyperparams}
Hyperparameters for Finetuning (batch size, learning rate, and finetuning epochs). ($^\star$internal dataset)}
\end{table}

\paragraph{Teacher Training} We follow the training procedure of XLM-Roberta \cite{conneau2020unsupervised}, and train the Base model (12 layers) on the CC100 dataset \cite{conneau2020unsupervised}. We use a peak learning rate of $6e-4$, the Adam optimizer with polynomial decay (with a warmup of 72K steps) and train the model for $1.5$ million steps, on a masked language modelling (MLM) objective with a masking probability of $0.15$.

\paragraph{Hidden State Distillation} We distill our KD-NAS recommended architectures on a subset of the CC100 dataset (with data in 116 languages) used to train the teacher. We train for $130,000$ steps, using an AdamW Optimizer \cite{loshchilov2019adamw} (with $\beta_1 = 0.9, \beta_2=0.98$). We use a peak learning rate of $8e-4$ with a linear LR schedule with a warmup of 6.5K steps.  We use a batch size of $32$ and train on 2 NVIDIA A100 GPUs. 

\paragraph{KD-NAS Pipeline}
Our KD-NAS controller is a single layer LSTM with $32$ cells, and is trained for 10 epochs per episode. We use a RMSProp optimizer \cite{HintonRMSProp} (with $\gamma=0.95$), a peak learning rate of $1e-4$ and an exponential learning rate schedule. For the proxy-set distillation, we follow the same parameters described above, except that we train for $4$ epochs on $30\%$ of the distillation data. We use a batch size of $32$ and distil on 1 NVIDIA A100 GPU.

\begin{table*}[ht!]
\centering
\begin{adjustbox}{max width=\textwidth}
\begin{tabular}{c|cccccccccc|c}
\hline
\textbf{Layer } & \multicolumn{10}{c|}{\textbf{GLUE}} & \textbf{XNLI} \\
 \textbf{Mapping}& \textbf{MNLI} & \textbf{QQP} & \textbf{QNLI} & \textbf{SST-2} & \textbf{CoLA} & \textbf{STS-B} & \textbf{MRPC} & \textbf{RTE} & \textbf{Avg.} & \textbf{Avg. (-CoLA)} & \textbf{Avg.} \\
 \hline
Last-1 & 77.6 & 85.9 & 85.2 & 88.5 & 25.1 & \textbf{84.8} & 86.2 & 57.3 & 73.8(0.7) & 80.8(0.6) & 56.2(0.4) \\
Last & 79.4 & 86.3 & 86.6 & 89.2 & 25.8 & 83.3& 85.1 & 62.1 & 74.7(1.0) & 81.7(0.9) & 63.1(0.4) \\
Uniform & 78.8 & 85.9 & 86.0 & 89.4 & 26.2 & 84.0 & 87.3 & 60.0 & 74.7(0.8) & 81.6(0.5) & 59.9(0.6) \\
Uniform+Last & \textbf{80.8} & \textbf{86.8} & \textbf{87.9} & \textbf{90.2} & \textbf{32.3} & 84.7 & \textbf{88.5} & \textbf{62.6} & \textbf{76.7(0.6)} & \textbf{83.1(0.3)} & \textbf{67.0(0.4)}\\
\hline
\end{tabular}
\end{adjustbox}
\caption{\label{tab:hs_distillation}
Layer Mapping Strategies for Hidden State KD for the 6L-768H Student. Average scores reported over 3 seeds (standard deviation in parenthesis). Best results in bold.}
\end{table*}

\begin{table*}[ht!]
\centering
\begin{adjustbox}{max width=\textwidth}
\begin{tabular}{c|cccccccc|cc}
\hline
\textbf{Model} & \multicolumn{10}{c}{\textbf{GLUE}} \\
 & \textbf{MNLI} & \textbf{QQP} & \textbf{QNLI} & \textbf{SST-2} & \textbf{CoLA} & \textbf{STS-B} & \textbf{MRPC} & \textbf{RTE} & \textbf{Avg.} & \textbf{Avg. (-CoLA)} \\
 \hline
KDNAS\textsubscript{Arch1} & 71.0 & 80.6 & 81.8 & 85.9 & 7.7 & 76.5 & 82.6 & 54.3 & 67.6(0.7) & 76.1(0.6) \\
KDNAS\textsubscript{Arch2} & 72.7 & 82.4 & 82.7 & 84.9 & 12.6 & 71.6 & 81.7 & 56.6 & 68.2(0.5) & 76.1(0.3) \\
KDNAS\textsubscript{Arch3} & 76.5 & 84.1 & 84.5 & 88.1 & 18.6 & 82 & 82.4 & 59.1 & 71.9(0.8) & 79.5(0.6) \\
\hline
6L-384H\textsuperscript{$\dagger$} & 76.1 & 83.7 & 83.8 & 87.5 & 17.8 & 78.2 & 84.4 & 57.3 & 71.1(1.0) & 78.7(0.8) \\
6L-768H\textsuperscript{$\dagger$} & 80.8 & 86.8 & 87.9 & 90.2 & 32.3 & 84.7 & 88.5 & 62.6 & 76.7(0.6) & 83.1(0.3) \\
DistilBERT & 78.8 & 85.9 & 86.9 & 88.9& 29.1 & 84.5 & 87.9 & 60.4 & 75.3(0.6) & 81.4(0.5) \\
\hline
Random\textsubscript{Seed1} &64.4&	72.4&	65.1&	85.2&	14.9&	17.6&	81.2&	52.2&	56.6(6.5) &	62.6(6.2) \\
Random\textsubscript{Seed2}& 74.3&	82.6&	83.3&	87.4&	16.9&	66.4&	83.7&	55.8&	68.7(3.2)&	76.1(3.8) \\
Random\textsubscript{Seed3}& 67.7&	73.8&	64.8&	86.1&	17.4&	36.5&	81.1&	55.2&	60.3(2.0)&	66.5(3.0) \\
\hline 
Teacher & 84.1  & 87.9& 90.2 & 91.9 & 51.7 & 86.6 & 91.4 & 61.4 & 80.6(0.3) & 84.8(0.3)\\
\hline
\end{tabular}
\end{adjustbox}
\caption{\label{tab:glue}
Performance on the GLUE Benchmark. Average taken over three seeds, standard deviation in parenthesis.}
\end{table*}

\begin{table*}[ht!]
\centering
\begin{adjustbox}{max width=\textwidth}
\begin{tabular}{c|ccccccccccccccc|c}
\hline
\textbf{Model} & \multicolumn{16}{c}{\textbf{XNLI}} \\
 & \textbf{ar} & \textbf{bg} & \textbf{de} & \textbf{el} & \textbf{en} & \textbf{es} & \textbf{fr} & \textbf{hi} & \textbf{ru} & \textbf{sw} & \textbf{th} & \textbf{tr} & \textbf{ur} & \textbf{vi} & \textbf{zh} & \textbf{Avg.} \\
 \hline
KDNAS\textsubscript{Arch1} & 55.0 & 61.6 & 60.8 & 59.5& 72.8 & 62.6 & 61.5 & 52.0 & 57.5 & 46.7 & 54.1 & 55.9 & 50.7 & 60.4 & 57.8 & 57.9 (0.5) \\
KDNAS\textsubscript{Arch2} & 55.9 & 62.1 & 61.2 & 61.9 & 73.2 & 62.7 & 62.0 & 52.7 & 57.2 & 45.9 & 57.3 & 54.9 & 48.5 & 59.9 & 58.9 & 58.3(0.6) \\
KDNAS\textsubscript{Arch3}& 58.8 & 64.8& 63.8 & 64.4 & 76.4 & 68.2 & 65.5 & 58.2 & 64.1 & 48.8 & 62.6 & 58.7 & 55.9 & 65.1 & 64.6 & 62.7(0.6) \\
\hline
6L-384H\textsuperscript{$\dagger$} & 59.7 & 67.2 & 63.4 & 65.6 & 75.9 & 68.7 & 66.8 & 58.3 & 62.4 & 48.9 & 62.7 & 59.1 & 53.4 & 63.2 & 65.1 & 62.7 (0.4) \\
6L-768\textsuperscript{$\dagger$} & 64.7 & 69.7 & 69.6 & 69.2 & 80.7 & 72.0 & 70.2 & 64.6 & 67.7 & 51.2 & 65.3 & 62.5 & 58.9 & 70.4 & 68.6 & 67.0 (0.4) \\
DistilBERT& 58.3&59.7 & 66.1 & 58.9 & 78.9 & 70.2 & 67.9 & 52.9 & 63.2 & 47.7 & 37.1 & 55.5 & 52.7 & 56.9 & 64.7 & 59.4 (0.5) \\
\hline
Random\textsubscript{Seed1} & 53.3	&59.9&	58.1&	57.8&	69	&58.8&	58&	51.5&	59.3&	43.5&	56&	53.9&	48.5&	57.9&	57.9&	56.2 (5.7) \\
Random\textsubscript{Seed2} & 56.5	&62	&64.2	&61.8	&75.9	&66.5	&65.1	&55.2	&62.1	&47.4	&61.1	&55.4	&52.0 &	63.8&	62.2&	60.6(5.3) \\
Random\textsubscript{Seed3} & 54.3&	60.4&	61.9&	59.0&	72.0&	61.2&	60.5&	54.2&	60.4&	43.7&	57.8&	55.6&	50.3&	58.9&	60.4 &	58.0(1.8) \\
\hline 
Teacher & 69.1 & 73.2 & 74.1 & 72.2 & 83.4 & 75.1 & 73.1 & 69 & 71.3 & 57.3 & 69.7 & 67.7 & 64.1 & 70.8 & 73.3 & 70.9 (0.8)\\
\hline
\end{tabular}
\end{adjustbox}
\caption{\label{tab:xnli}
Performance on the XNLI Task. Average taken over three seeds, standard deviation in parenthesis.}
\end{table*}

\paragraph{Finetuning Hyperparameters}
To finetune our models, we did a coarse grid search on learning rate \{$1e-5, 2e-5, 3e-5, 5e-5$\} and number of finetuning epochs \{$3,5,10$\} for a few models, and found that the hyperparameters shown in \cref{tab:hyperparams} generally work well across models. However, for our internal NER dataset, the teacher model requires $6$ epochs with a learning rate of $5e-6$. We also find finetuning on CoLA for a longer number of epochs particularly helpful for smaller models.

\section{Text Classification Results}\label{app:results}
We show performance of our KD-NAS models and Baselines on all tasks of the GLUE Benchmark, comprising of English text classification tasks in \cref{tab:glue}. We show the per-language performance of the models on XNLI, a cross lingual inference task, trained on English and evaluated on 15 languages in \cref{tab:xnli}.

\section{Internal Named Entity Recognition Dataset}\label{app:wnlpner}
\begin{figure}[t!]
\includegraphics[width=0.5\textwidth]{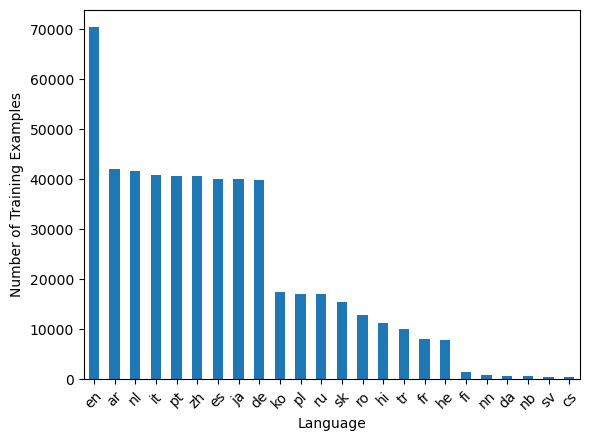}
\caption{Per-language distribution of training examples for our internal multilingual NER dataset. }
\label{fig-wnlp-train-ex}
\end{figure}

\begin{figure}[t!]
\includegraphics[width=0.5\textwidth]{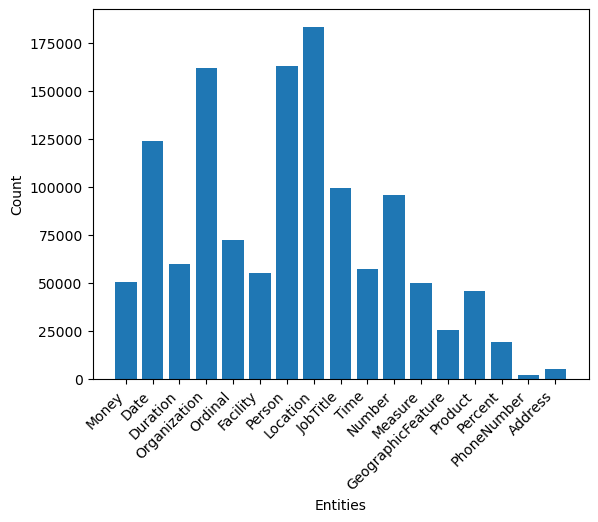}
\caption{Distribution of entities in training set for our internal multilingual NER dataset.}
\label{fig-wnlp-entities}
\end{figure}

We use an internal multilingual Named Entity Recognition Dataset to evaluate our models, which contains of $17$ entities across $24$ languages. This set comprises of training set of 518K examples, a development set of 54K examples and a test set of 59K examples. The per-language distribution of training examples is shown in \cref{fig-wnlp-train-ex}. On average, each training example consists of around 79 words, and about 9 entity mention spans. We show the distribution of these entities across the training set in \cref{fig-wnlp-entities}.

\begin{table*}[t!]
\centering
\begin{adjustbox}{max width=\textwidth}
\begin{tabular}{c|c|ccc|ccc|c}
\hline
\textbf{Model} & \textbf{Distillation}  & \multicolumn{3}{c|}{\textbf{Monolingual Performance} }  & \multicolumn{3}{c|}{\textbf{Multilingual Performance} }& {\textbf{Average} } \\
 & \textbf{Algorithm} &  GLUE & SQuAD & CoNLL & XNLI & TyDi & NER\textsuperscript{$\star$} &   \\
 \hline
KDNAS\textsubscript{Arch3} & HS Transfer & 79.5(0.6) & 62.7(0.1) & 92.1(0.2) & 62.7(0.5) & 66.8(0.6) & 72.8(0.06) & 72.8 \\
KDNAS\textsubscript{Arch3} & MiniLM & 80.7(0.9) & 67.4(0.5) & 92.5(0.1) & 65.3(0.8) & 68.6(0.3) & 74.2(0.05) & 74.8 \\
\hline
Teacher &  & 84.8(0.3) & 77.3(0.5) & 94.2 (0.1) & 70.9 (0.8) & 78.1(0.7) & 76.1(0.06) & 80.2\\
\hline
\end{tabular}
\end{adjustbox}
\caption{\label{tab:piccolo-perf}
Performance of distilling the Piccolo Model (KDNAS\textsubscript{Arch3}). Avg. GLUE score across all tasks (except CoLA). Avg XNLI accuracy for all languages. F1 score reported for SQuAD, TyDi and CoNLL. Results shown as an average over 3 seeds, with the average standard deviation shown in parenthesis. ( $^\star$internal dataset)}
\end{table*}
\section{MiniLMv2 Distillation Performance}\label{app:minilmv2-perf}

MiniLM v2 \cite{wang-etal-2021-minilmv2} proposes to train the student model to match the self attention relations of the teacher. Formally, let $d_h$ denote the hidden size, and $A_h$ denote the number of attention heads of a transformer with $L$ layers, and $|x|$ be the sequence length. We first individually concatenate each of the Query, Key and Value vectors $\mathbf{Q}_{i,a}, \mathbf{K}_{i,a}, \mathbf{V}_{i,a} \in \mathbb{R}^{|x| \times \frac{d_h}{A_h}}$ $(a \in [1, A_h])$ for layer $i$. They are then split into $A_r$ relation heads $\mathbf{A}_{Q,i,r}, \mathbf{A}_{K,i,r}, \mathbf{A}_{V,i,r} \in \mathbb{R}^{|x| \times d_r}$ ($r \in [1, A_r]$ and $d_r = \frac{d_h}{A_r}$). The \textit{self-attention relations} are the scaled dot products of these matrices, or
\begin{equation}
    \textbf{R}_{\alpha\beta, i, r} = \mathrm{softmax}\Bigl( \frac{\mathbf{A}_{\alpha,l,r}\cdot \mathbf{A}_{\beta,l,r}^{\mathsf{\,T}}}{\sqrt{d_r}} \Bigl)
\end{equation}
Where $\alpha, \beta \in \{Q, K, V\}$. The distillation objective minimizes the cross entropy loss between the teacher and student self attention relations, $\textbf{R}_{\alpha\beta, j, r}^T$ and $\textbf{R}_{\alpha\beta, i, r}^S$:
\begin{equation}
\mathcal{L}_{\mathrm{MiniLM}} = \!\! \sum_{\alpha,\beta \in \{ Q,K,V \}} \sum_{r = 1}^{A_r} \mathrm{CE} \Bigl( \mathbf{R}^T_{\alpha, j, r},  \mathbf{R}^S_{\alpha, i, r} \Bigl)
\label{eq:mha_transfer}
\end{equation}

Since $ \textbf{R}_{\alpha\beta, i, r} \in \mathbb{R}^{|x| \times |x|}$, the distillation objective is architecture-agnostic and allows for the teacher and student to have different hidden sizes and number of attention heads. 

We distill only the Q-Q, K-K, and V-V self attention relations (i.e., $\alpha = \beta$) of the $11^{th}$ teacher layer into the last student layer. We find that distilling the second-last teacher layer gives the best performance (in line with the observations of \citet{wang-etal-2021-minilmv2}), and we use $48$ relation heads. Our deployed model was distilled on the entire CC100 corpus for $200,000$ steps, using a batch size of $16$ on 30 NVIDIA V100 GPUs.

We show the performance improvement by distilling the Piccolo Model (KDNAS\textsubscript{Arch3}) on the task-agnostic, self-attention relation distillation objective of MiniLM-v2 in \cref{tab:piccolo-perf}. Due to the improvement of performance on all tasks, we use this model for final deployment. We aim to perform the KD-NAS process with a MiniLM-based distillation objective as future work.

\end{document}